\documentclass[journal]{IEEEtran}

\ifCLASSINFOpdf

\else
 
\fi

\usepackage{array}
\usepackage{amsmath}
\usepackage{graphicx}
\usepackage{upgreek}
\usepackage{multirow}
\usepackage{color}
\usepackage{pifont}
\usepackage{url}
\usepackage{bm}
\usepackage{threeparttable}
\usepackage{float}
\usepackage{enumitem}
\usepackage{amssymb}
\usepackage{booktabs}   

\usepackage[colorlinks,linkcolor=blue]{hyperref}

\begin{document}
 
\title{Regression in EO: Are VLMs Up to the Challenge?}

\author{Xizhe~Xue, Xiao Xiang~Zhu*

\IEEEcompsocitemizethanks{

Xizhe Xue and Xiao Xiang Zhu are with the Chair of Data Science in Earth Observation, Technical University of Munich, 80333 Munich, Germany (e-mails: xizhe.xue@tum.de; xiaoxiang.zhu@tum.de)

Xiao Xiang Zhu is also with Munich Center for Machine Learning, 80333 Munich, German.
} 
}


\maketitle

\section*{Abstract}

Earth Observation (EO) data encompass a vast range of remotely sensed information, featuring multi-sensor and multi-temporal, playing an indispensable role in understanding our planet’s dynamics. Recently, Vision Language Models (VLMs) have achieved remarkable success in perception and reasoning tasks, bringing new insights and opportunities to the EO field. However, the potential for EO applications, especially for scientific regression related applications remains largely unexplored. This paper bridges that gap by systematically examining the challenges and opportunities of adapting VLMs for EO regression tasks. The discussion first contrasts the distinctive properties of EO data with conventional computer vision datasets, then identifies four core obstacles in applying VLMs to EO regression: 1) the absence of dedicated benchmarks, 2) the discrete-versus-continuous representation mismatch, 3) cumulative error accumulation, and 4) the suboptimal nature of text-centric training objectives for numerical tasks. Next, a series of methodological insights and potential subtle pitfalls are explored. Lastly, we offer some promising future directions for designing robust, domain-aware solutions. Our findings highlight the promise of VLMs for scientific regression in EO, setting the stage for more precise and interpretable modeling of critical environmental processes.

\begin{figure*}
    \centering
    \includegraphics[width=1.0\linewidth]{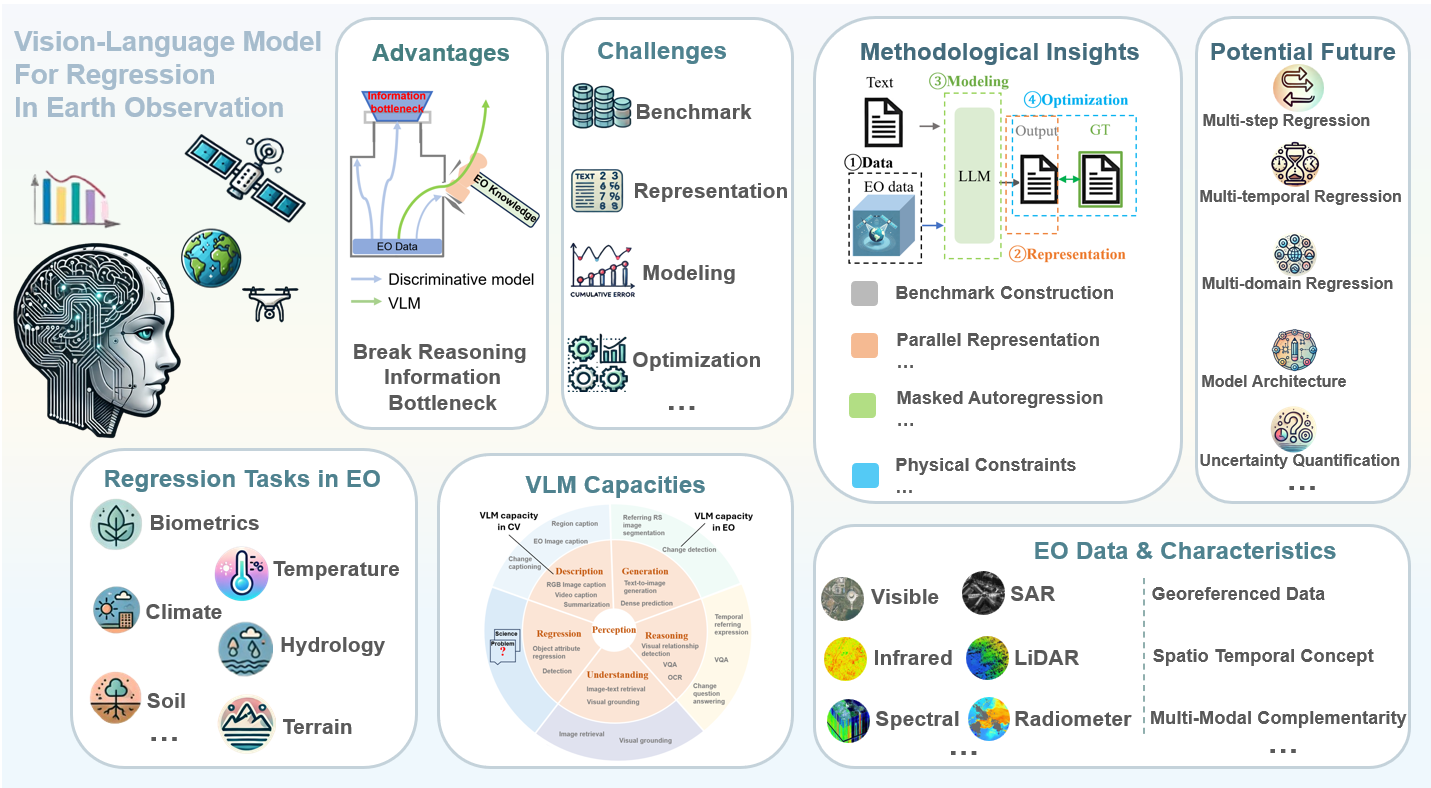}
    \caption{Structure and rationale of this paper. Based on an analysis of regression tasks, data characteristics in EO, and  VLM capabilities, we investigate the advantages, challenges, potential solutions, and promising future directions for utilizing VLMs in regression tasks within EO.}
    \label{fig:abstract}
    \vspace{-2mm}
\end{figure*}

\section{Introduction}


Earth Observation (EO) ~\cite{tuia2023artificial} data serve as an indispensable resource for monitoring and analyzing Earth’s ever-changing environments. From satellite-based thermal imaging to high-resolution multispectral and radar data, these sensing modalities yield detailed insights into physical phenomena such as temperature variations, vegetation health, and atmospheric composition change. Therefore, Earth observation data are widely applied in various scientific problems and practical scenarios, such as global biodiversity monitoring \cite{skidmore2021priority}, land cover mapping \cite{brown2022dynamic}, and above ground biomass estimation \cite{li2020forest}. 

Among these applications, regression analysis of observation data based on regression techniques is of great importance. For instance, predicting soil moisture levels aids agricultural planning and irrigation management, mapping land surface temperatures supports climate research and urban heat island studies, and estimating snow-water equivalents is crucial for hydrological modeling and water resource management.

However, the interpretation of EO data has primarily relied on traditional image processing techniques, such as object detection and segmentation focused on specific targets or land cover classifications. These methods primarily utilize the visual information inherent in the imagery, without fully integrating the vast amounts of domain-specific knowledge available in the field. Consequently, the analysis is constrained by an information bottleneck, limiting the potential accuracy and depth of the inferred regression outcomes.

Recently, the remarkable advancements in large Vision Language Models (VLMs) demonstrated by their success in image captioning, visual question answering, and other perception-driven benchmarks~\cite{liu2023llava, gpt4}. As illustrated in Fig.~\ref{fig:abstract}, VLMs are fundamentally perceptual, with core capabilities encompassing description, understanding, reasoning, generation, and regression. Leveraging these abilities, VLMs can perform a wide array of tasks to interpret both CV and EO data.  These raise an interesting question: \emph{Are VLMs up to the regression tasks in EO?} 
To explore this overlooked problem, this perspective paper offers a foundational exploration into the application of VLMs for scientific regression tasks in EO. The main contributions of this work are:

\begin{enumerate} 

\item \textbf{Initial exploration of VLMs for EO Regression.}
This paper pioneers the investigation of VLMs' potential in addressing regression tasks within EO, specifically focusing on overcoming the information bottleneck that arises from relying solely on image modality for inference.

\item \textbf{Comprehensive analysis of existing challenges.}  
A multifaceted examination of the challenges faced by VLMs in EO regression is conducted, highlighting issues such as benchmark gaps, representation mismatches, error amplification, and suboptimal training objectives. These challenges, though significant, have been largely overlooked in prior research, underscoring the need for targeted solutions.

\item \textbf{Proposed methodological insights.}  
For each identified challenge, potential solutions are proposed, including parallel representation strategies, discrete-guided continuous prediction mechanisms, masked autoregression techniques, and optimization strategies tailored for numerical accuracy. These insights aim to lay the groundwork for future research and development in VLM-based EO regression.

\item \textbf{Identification of pitfalls and future research directions.}  
Subtle pitfalls, such as the loss of key visual details and scale variability across geographical regions, are identified and discussed. Additionally, promising future research directions are suggested, including the integration of multi-step numerical reasoning, handling multi-domain tasks, and incorporating uncertainty quantification to enhance the robustness and precision of EO regression models.
\end{enumerate}



This work establish a foundation for future VLM approaches that can satisfy the precision and interpretability requirements of diverse EO regression applications. The remainder of this paper is structured as follows. Section II  provides some fundamental knowledge and surveys related work. Section III offers the mathematical definition and distinctive features of the regression problem in the EO field. Section IV examines the challenges of adapting VLMs to EO regression.  Section V proposes a set of methodological insights and Section VI discusses potential pitfalls that can undermine model accuracy. Section VII  explores future research directions for robust EO regression.
Section VIII concludes the paper.


\begin{table*}[htbp]
\centering
\caption{A Summarized Comparison of EO Data and General CV Data}
\label{tab:eo_vs_cv}
\begin{tabular}{p{1.5cm}|p{7.5cm}|p{7.2cm}}
\hline
\textbf{Aspect} & \textbf{EO Data} & \textbf{General CV Data} \\
\hline
\textbf{Imaging Modalities} &
\begin{itemize}[leftmargin=*]
    \item Acquired via satellites, drones, aircraft
    \item Sensors: multispectral, hyperspectral, SAR, LiDAR
    \item Requires geometric/radiometric corrections
\end{itemize} &
\begin{itemize}[leftmargin=*]
    \item Predominantly RGB cameras (phones, surveillance)
    \item Occasionally RGB-D or thermal sensors
    \item Distortions mainly from lighting, motion blur
\end{itemize}
\\
\hline
\textbf{Spatial Resolution} &
\begin{itemize}[leftmargin=*]
    \item Varies greatly: tens of meters to sub-meter
    \item Different satellites / sensors yield different resolutions
\end{itemize} &
\begin{itemize}[leftmargin=*]
    \item Typically in the range of hundreds to a few thousand pixels
    \item Less variability than EO in a single dataset
\end{itemize}
\\
\hline
\textbf{Spectral Resolution} &
\begin{itemize}[leftmargin=*]
    \item Multiple bands
    \item SAR (microwave), LiDAR (distance/range) 
\end{itemize} &
\begin{itemize}[leftmargin=*]
    \item Mainly limited to visible RGB channels
    \item Some datasets include depth or thermal, but rarer
\end{itemize}
\\
\hline
\textbf{Temporal Characteristics} &
\begin{itemize}[leftmargin=*]
    \item Regular revisit (e.g., every 16 days)
    \item Long-term monitoring of slow changes (urban, vegetation)
\end{itemize} &
\begin{itemize}[leftmargin=*]
    \item Often single images or short videos
    \item Focus on events in seconds/minutes (tracking, actions)
\end{itemize}
\\
\hline
\textbf{Physical interpretability} &
\begin{itemize}[leftmargin=*]
    \item Pixel values can represent reflectance/backscatter
    \item Geo-referenced, each pixel has real-world coordinates
\end{itemize} &
\begin{itemize}[leftmargin=*]
    \item RGB values are not strictly physical measures
    \item No inherent geographic referencing
\end{itemize}
\\
\hline
\textbf{Labeling Complexity} &
\begin{itemize}[leftmargin=*]
    \item Requires domain expertise (remote sensing, geoscience)
    \item Change detection labeling across multiple time points
    \item Data can be huge, but labeled samples are scarce
\end{itemize} &
\begin{itemize}[leftmargin=*]
    \item Well-established labeling (ImageNet, COCO, etc.)
    \item Annotations often based on straightforward human vision
\end{itemize}
\\
\hline
\textbf{Data Scale} &
\begin{itemize}[leftmargin=*]
    \item Potentially TB--PB scale daily on a global basis
    \item Often processed in distributed/cloud environments
\end{itemize} &
\begin{itemize}[leftmargin=*]
    \item Large datasets too, but with mature data pipelines
    \item Real-time CV solutions well-established (e.g., surveillance)
\end{itemize}
\\
\hline
\textbf{Domain Adaptation} &
\begin{itemize}[leftmargin=*]
    \item Huge domain gaps (varying sensors, seasons, angles)
    \item Complex domain adaptation or transfer learning needed
\end{itemize} &
\begin{itemize}[leftmargin=*]
    \item Differences exist (ImageNet vs.\ COCO) but smaller gaps
    \item Rich pretrained models for easy transfer learning
\end{itemize}
\\
\hline
\textbf{Application Domains} &
\begin{itemize}[leftmargin=*]
    \item Large-scale: land cover, disaster monitoring, climate studies
    \item Results guide policy, economic strategies, emergency response
\end{itemize} &
\begin{itemize}[leftmargin=*]
    \item Human/object-centric tasks: face recognition, autonomous driving
    \item Focus on near-field scenarios, shorter time scale
\end{itemize}
\\
\hline
\end{tabular}
\end{table*}

\section{Fundamental knowledge}
\subsection{Earth Observation Data vs. General Computer Vision Data}
While EO data and general CV imagery both utilize digital imaging principles, they exhibit profound differences that critically impact model design. As summarized in Table~\ref{tab:eo_vs_cv}, these distinctions span several critical aspects: imaging modalities, spatial/spectral resolutions, temporal characteristics, physical interpretability, labeling protocols, data scale, domain adaptation requirements and applications, etc. Rather than simply listing technical contrasts, we focus on how these inherent divergences demand specialized architectural considerations for EO tasks.

\subsubsection{Modeling Challenges from Sensor Heterogeneity}
EO systems integrate diverse sensors (SAR, hyperspectral, LiDAR, etc.), creating multimodal fusion challenges rarely seen in CV. While CV models handle standardized RGB data \cite{he2016deep}, EO models must unify physically distinct signals like microwave backscatter and optical reflectance. Studies show that directly fusing SAR and optical data reduces model accuracy by 23-41\% \cite{zhang2022multimodal}. Promising solutions may focus on: (1)  Developing cross-sensor attention mechanisms; (2) Implementing physics-guided normalization protocols to maintain spectral fidelity during fusion.

\subsubsection{Temporal-Spatial Constraints in Learning}
EO's fixed revisit cycles (e.g., 16-day intervals) impose strict spatiotemporal constraints. Incorporating orbital parameters boosts crop prediction accuracy by 18.7\% \cite{chen2023orbital}, outperforming generic sequence models. Multi-temporal analysis requires sub-pixel alignment accuracy below 0.3 pixels \cite{lee2022registration}. This precision demand necessitates specialized geometric correction modules that go beyond traditional CV methods primarily designed for single-image processing.

\subsubsection{Physically-Grounded Semantic Learning}
EO pixels represent measurable biophysical properties (NDVI, soil moisture), unlike CV's visual features \cite{zeiler2014visualizing}. Ignoring physical laws causes error amplification (1.7dB/stage \cite{reichstein2019deep}). To mitigate this, recent advances suggest two synergistic strategies: (1) Embedding differentiable physics models as trainable regularizers \cite{willard2020integrating}; (2) Enforcing interpretability constraints between low-level measurements and domain-specific semantics. These approaches aim to suppress nonphysical correlations that conventional CV architectures might inadvertently learn.

\subsection{Regression Tasks in Earth Observation}

Regression tasks in EO focus on estimating continuous scientific variables through multimodal remote sensing data. These quantitative approaches underpin critical scientific discoveries and operational monitoring systems. We systematically review representative regression applications across seven key domains: biometrics, climate, soil, temperature, hydrology, terrain, and other emerging frontiers, highlighting their unique methodological requirements.

\subsubsection{Biometrics: Vegetation and Biomass Estimation}
This category \cite{liu2020global,saatchi2011benchmark} involves quantifying plant community structure, density, and total organic matter using EO data such as LiDAR, radar, and multispectral or hyperspectral imagery. Accurate estimations of vegetation and biomass are crucial for ecosystem health assessments, carbon stock monitoring, and environmental change studies. They directly support forestry management, climate modeling, and conservation planning.

\subsubsection{Climate: Atmospheric Parameter Estimation}
Under this heading, models predict atmospheric properties such as humidity, gas concentrations, and related parameters from thermal infrared satellite sensors (e.g., MODIS, Landsat TIRS) and spaceborne instruments (e.g., TROPOMI)~\cite{li2024vision}. Representative applications \cite{eldering2017oco,saatchi2011benchmark} include land surface temperature forecasting, greenhouse gas monitoring, and air quality analysis. Accurate estimation of these parameters is essential for understanding and managing climatic and environmental health.

\subsubsection{Soil: Land and Soil Parameter Estimation}
EO-based regression methods measure and analyze soil characteristics such as moisture content and organic carbon using microwave (Sentinel-1, SMAP) and hyperspectral data~\cite{entekhabi2014smap,vreugdenhil2020soil}. These estimates inform agricultural irrigation practices, flood prediction refinement, carbon cycle investigations, and environmental management.

\subsubsection{Temperature: Land Surface and Marine Temperature}
This category \cite{donner2020machine, wan2014new} includes tasks that model surface temperature or sea surface temperature (SST) via thermal (AVHRR, MODIS, VIIRS) and in-situ observations~\cite{reichstein2019deep}. Example applications include investigating urban heat islands, forecasting weather patterns, and studying land-atmosphere interactions.

\subsubsection{Hydrology: Water and Flood-Related Parameter Estimation}
These applications involve determining aquatic metrics such as water quality indices, sea surface temperature, and flood inundation levels using thermal and multi/hyperspectral sensors \cite{ampatzidis2020global} . Flood depth and extent estimates from time-series SAR (e.g., Sentinel-1) \cite{tsyganskaya2018flood} inform disaster management and damage assessments. Meanwhile, surface roughness and soil moisture estimations are critical for hydrological modeling, water resource planning, and agriculture.

\subsubsection{Terrain: Topographic and Geological Parameter Estimation}
This group of tasks  \cite{farr2007shuttle,zink2014tanDEM} . focus on modeling the Earth's surface features, including elevation, slope, and geological attributes~\cite{zhang2024earthgpt}. LiDAR and radar interferometry (InSAR) datasets are processed with regression or interpolation methods to produce high-resolution Digital Elevation Models (DEMs). Such DEMs underpin flood modeling, infrastructure planning, geological assessments, and soil erosion studies.

\subsubsection{Other Emerging Frontiers: Urban, Socioeconomic, and Multipurpose Applications}
\begin{itemize}
    \item \textbf{Urban Heat Island Analysis:} Thermal infrared data and regression models quantify intra-urban temperature variations, assisting urban planners in mitigating heat islands \cite{li2020influence}.
    \item \textbf{Building Height Estimation:} Shadows in optical or SAR imagery can be converted into building height estimates, supporting 3D city modeling and urban planning\cite{fan2024pano2geo}.
    \item \textbf{Population Density Prediction:} Nightlight (DMSP/OLS, VIIRS/DNB) and high-resolution optical imagery are integrated via regression to derive population density maps. Accurate population mapping is essential for resource allocation and crisis management \cite{boo2022high}.
    \item \textbf{Agriculture and Crop Monitoring:} Time-series satellite data (MODIS, Sentinel-2) and climate variables enable yield prediction, acreage estimation, and crop health assessments\cite{tsyganskaya2018flood}. 
    \item \textbf{Snow and Ice Monitoring:} Multi-sensor data (microwave, optical) and regression models measure snow water equivalent (SWE) and sea ice thickness, vital for water resource planning and polar studies \cite{entekhabi2014smap}.
    \item \textbf{Solar Radiation Estimation:} Regression techniques linking satellite measurements to ground observations provide surface solar irradiance predictions for renewable energy planning~\cite{qi2024mapping}.
    \item \textbf{Socioeconomic Indicators:} Integrating nightlight data, urban footprints, and demographic records helps estimate economic activity or poverty indices, guiding policy decisions and infrastructure planning\cite{zhang2016socioeconomic}.
\end{itemize}

Overall, these regression tasks in EO span an extensive range of environmental, climatic, and socioeconomic applications. Ongoing advances in machine learning, data fusion, and sensor integration promise even higher accuracy, finer spatial resolution, and richer insights into our evolving earth system.

\subsection{Large Language Models for Earth Observation}
In this section, we categorize early influential Transformer-based~\cite{vaswani2017attention} LLMs into three primary types based on their neural architectures: encoder-only, decoder-only, and encoder-decoder models. For a detailed overview of early LLM developments, refer to~\cite{zhao2023survey,minaee2024large}. Following this classification, we examine several LLMs relevant to EO.
\subsubsection{Encoder-only LLMs}  Encoder-only models are comprised solely of an encoder network. These models were initially designed for tasks requiring language comprehension, such as text classification, where the objective is to predict class labels based on input text. Prominent examples in this category include BERT~\cite{devlin2018bert} and its derivatives, such as RoBERTa~\cite{liu2021robustly}, ALBERT~\cite{lan2019albert}, DeBERTa~\cite{he2020deberta}, XLM~\cite{conneau2019cross}, XLNet~\cite{yang2019xlnet}, and UNILM~\cite{dong2019unified}.
\subsubsection{Decoder-only LLMs} Two notable families of decoder-only models are the GPT series and the LLaMA~\cite{LLaMA} family. The GPT models, starting with GPT-1~\cite{gpt1} and GPT-2~\cite{gpt2}, established the foundation for more advanced models like GPT-3~\cite{gpt3} and GPT-4~\cite{gpt4}. LLaMA, developed by Meta, offers a series of open-source foundational models. While LLaMA shares the transformer architecture with GPT-3, it incorporates minor architectural adjustments, distinguishing it from GPT models.
\subsubsection{Encoder-Decoder LLMs} As demonstrated by Raffel et al. in ~\cite{raffel2020exploring}, most NLP tasks can be reformulated as sequence-to-sequence generation problems. Encoder-decoder models are inherently versatile, enabling them to handle both natural language understanding and generation tasks. Noteworthy models in this category include T5~\cite{raffel2020exploring}, mT5~\cite{xue2020mt5}, MASS~\cite{song1905mass}, and BART~\cite{lewis2019bart}.
\subsubsection{Domain-Specific LLMs for Geoscience and EO}
LLMs tailored for geoscience and EO  handle domain-specific tasks by enabling more accurate information retrieval, scientific text understanding, and quantitative analysis. Unlike general-purpose LLMs, these models are fine-tuned on geoscience corpora, allowing them to interpret technical terminology, synthesize cross-disciplinary knowledge. Several notable domain-specific LLMs have been developed to address different challenges within geoscience and EO. Here, we  review several representative models and their contributions to this field.

GeoBERT~\cite{gao2022geobert} is a BERT-based model trained on 20 million geoscience records. It is optimized for scientific text mining, question answering, and summarization. Its strong domain adaptation enhances semantic understanding in geoscientific literature, aiding researchers in information extraction and literature analysis.

OceanGPT~\cite{bi2023oceangpt} is designed for ocean science applications, leveraging the DoInstruct framework to generate domain-specific instructions. It excels in ecosystem-related regression tasks, such as water temperature prediction and marine biomass estimation, addressing the complexities of oceanographic data.

GeoGalactica~\cite{lin2023geogalactica} represents one of the largest geoscience-specific LLMs, integrating instruction-tuned datasets to support tasks like mineral potential prediction and climate analysis. Its ability to synthesize information across different geoscience subfields makes it a powerful tool for quantitative research.

CnGeoPLM~\cite{ma2023cngeoplm} is a Chinese geological language model optimized for geological text processing. It specializes in entity recognition, relationship extraction, and geological structure prediction, enabling efficient analysis of Chinese-language geoscientific data.

These models exemplify how domain-specific LLMs enhance geoscience research. As geoscience datasets continue to grow, further advancements in model architectures and fine-tuning strategies will be crucial for enhancing their interpretability and application scope.

\subsection{Large Language Models for Regression Tasks}
In recent years, LLMs have demonstrated exceptional capabilities in handling large-scale textual data, extracting meaningful patterns from vast and diverse datasets. Their success extends beyond conventional natural language processing tasks, finding applications in fields such as coding~\cite{li2022competition}, symbolic mathematics~\cite{lewkowycz2022solving}, and scientific reasoning~\cite{singhal2022large}. Given these versatile applications, a natural question arises: \textbf{Can LLMs be effectively applied to regression tasks}, where the goal is to predict continuous numerical outputs based on input features? This intriguing possibility has sparked research into adapting LLMs for numerical prediction tasks, opening new avenues for exploration in data-driven modeling. To provide a comprehensive overview of foundational concepts, we review several representative works that highlight key directions and advancements in applying LLMs for regression tasks.
\subsubsection{Open-Source Models: Training from Scratch vs. Fine-Tuning}

Encoder-decoder models have been widely adopted in various NLP tasks, including numerical regression in EO applications. These models leverage both encoding and decoding mechanisms, making them inherently suitable for sequence-to-sequence learning. However, depending on the training paradigm, they can be developed either from scratch or through fine-tuning pre-trained models. Two representative encoder-decoder approaches are discussed here to illustrate these contrasting strategies.

\paragraph{Training from Scratch}  
OMNIPRED~\cite{song2024omnipred} introduces an end-to-end framework that trains LLMs as universal regressors from scratch. Built on a 200M parameter T5 encoder-decoder architecture~\cite{raffel2020exploring}, it formulates numerical regression as a sequence prediction problem. The model encodes input features in a structured key-value format and represents numerical outputs with fixed-length tokens to ensure precision. Through full-model training with cross-entropy loss on numerical tokens, OMNIPRED achieves accurate predictions across diverse regression tasks without relying on pre-existing language models.

\paragraph{Fine-Tuning a Pre-Trained Model}  
The Embed-Then-Regress framework~\cite{nguyen2024} fine-tunes only the regression-specific layers of a pre-trained T5-XL encoder-decoder model. Instead of training from scratch, it generates compact fixed-dimensional embeddings from tokenized inputs, allowing traditional regression models to operate efficiently. This approach pools token embeddings to construct a condensed representation, enabling effective numerical regression while significantly reducing computational overhead.

\subsubsection{Closed-Source Models: Decoder-Only In-Context Learning}  

Recent studies~\cite{vacareanu2024words} have demonstrated that closed-source, decoder-only LLMs, such as GPT-4~\cite{gpt-4} and LLaMA2~\cite{touvron2023llama}, exhibit strong regression capabilities through in-context learning (ICL). Despite being originally designed for next-token prediction, these models can perform numerical regression by leveraging in-context exemplars, bypassing the need for task-specific fine-tuning. This emergent capability highlights the adaptability of decoder-only architectures for regression-based EO tasks.

\subsubsection{Theoretical Insights: Understanding LLM Embeddings for Regression}  

Theoretical investigations~\cite{tang2024understanding} have explored how LLM embeddings function in numerical regression tasks. One notable concept, the Normalized Lipschitz Factor Distribution (NLFD)~\cite{virmaux2018lipschitz}, quantifies embedding smoothness and sensitivity to input perturbations. Findings suggest that models with highly skewed NLFDs tend to underperform in regression, indicating that embedding stability is a key determinant of numerical prediction accuracy. Interestingly, model size and general language pre-training did not consistently correlate with better regression performance, underscoring the need for specialized adaptation techniques.

Despite these advancements, several open challenges remain. The impact of different LLM architectures (encoder-only, encoder-decoder, and decoder-only), training methodologies, optimization strategies, and post-processing techniques on regression accuracy requires further exploration. 

\subsection{Vision Language Models for Earth Observation}
Recent advancements in VLMs have significantly improved EO applications by enhancing multi-modal data analysis~\cite{li2024vision}. These models can be broadly categorized into two groups: contrastive VLMs, which emphasize vision-text alignment through contrastive learning, and conversational VLMs, which integrate large language models (LLMs) to facilitate interactive and instruction-based geospatial understanding.

\subsubsection{Contrastive VLMs for EO}

Contrastive VLMs~\cite{radford2021learning} primarily consist of an image encoder and a text encoder, with the fundamental objective being the precise alignment of extracted features from both modalities. Typically, the image encoder based on CNN or ViT architectures  converts images into latent feature representations, while the text encoder maps textual descriptions into corresponding embeddings. The model is trained using a contrastive loss, such as InfoNCE, to maximize the similarity of semantically related image-text pairs while minimizing irrelevant pair associations.

CLIP has pioneered this vision-language pretraining strategy, laying the foundation for its adaptation to remote sensing. Recent research focuses on transferring the strong alignment capabilities developed in natural images to remote sensing data, thereby enabling diverse downstream tasks, including change detection and remote sensing image captioning. One of the earliest remote sensing-specific contrastive VLMs, RemoteCLIP~\cite{remoteclip}, introduced by Liu \textit{et al.}, refines CLIP for the geospatial domain. To address the challenge of limited annotated datasets, RemoteCLIP is trained on a curated image-caption dataset using continual learning, ensuring robust semantic feature extraction and efficient text alignment.

Following a similar methodology, GeoRSCLIP \cite{rs5mgeorscliplargescale} constructs a large-scale remote sensing dataset containing five million image-text pairs by filtering existing datasets and generating textual descriptions for previously caption-less images. Fine-tuning CLIP on this dataset enhances its adaptability to remote sensing tasks. In contrast, CRSR \cite{cross-modal} introduces a Transformer Mapper network with attention-based semantic query embeddings, offering a more refined representation of remote sensing imagery. This structure enhances the model’s ability to discern subtle variations and extract richer semantic information.

Several other approaches extend CLIP's capabilities by refining the alignment process. ProGEO \cite{progeo} augments geographic image pretraining by incorporating additional descriptive text, which refines the image encoder’s capacity to capture detailed features. This enhancement significantly improves geo-localization tasks by aligning spatial characteristics with textual descriptors. CRAFT \cite{RemoteSV} takes a different approach by aligning satellite images with their co-located internet image embeddings, thereby eliminating the need for text annotations during training. ChangeCLIP \cite{changeclip} innovates by introducing a DFC layer, allowing the model to learn transformation-specific features in bitemporal remote sensing images rather than focusing solely on semantic categories.

Additional research efforts improve CLIP-based models by optimizing pseudo-label strategies. S-CLIP~\cite{sclip} enhances CLIP’s generalization ability by categorizing pseudo-labels into title-level and keyword-level descriptions, leveraging their complementary advantages. Meanwhile, RS-CLIP \cite{rs-clip} refines this strategy by autonomously generating pseudo-labels from unlabeled data and incorporating a curriculum learning framework to iteratively refine the training process.

\subsubsection{Conversational VLMs for EO}

Conversational VLMs~\cite{LLaVA} have emerged as powerful tools for tackling complex EO tasks, leveraging LLMs~\cite{LLaMA} to transform high-dimensional visual data into structured textual insights. By integrating advanced image encoders that capture fine-grained geospatial features with text encoders optimized for understanding natural language queries, these models demonstrate remarkable performance, particularly in few-shot learning and interactive scenarios.

A critical component of conversational VLMs is the \textbf{alignment layer}, which bridges remote sensing imagery and LLMs. Common techniques involve using Multi-Layer Perceptrons (MLPs) for feature projection and incorporating learnable query embeddings to extract task-specific image representations. By leveraging pre-trained LLM, these models first process images through a vision encoder and then translate the extracted features into a format that an LLM can interpret, enabling a wide range of vision-language tasks. Various models adopt different strategies to optimize vision-language alignment.

\textbf{Image encoder} severs as a key and fundemental part in VLMs.  While the majority of approaches rely on the standard CLIP image encoder \cite{clip}, some models have explored alternative designs to enhance performance. For instance, RSGPT \cite{rsgpt} and SkyEyeGPT \cite{skyeyegpt} incorporate the EVA-CLIP encoder \cite{eva-clip}, which benefits from improved training methodologies, leading to better feature extraction and representation learning. Meanwhile, RS-LLaVA \cite{RS-llava} modifies the LLaVA \cite{LLaVA} framework for remote sensing by introducing a domain-specific image encoder trained on specialized datasets. Through instruction tuning and precise alignment, RS-LLaVA effectively captures geospatial semantics, producing contextually relevant responses tailored to remote sensing applications. EarthGPT~\cite{earthgpt} employs a Vision Transformer image encoder to extract hierarchical visual features, complemented by a CNN backbone that integrates multi-scale local information. This multi-layered feature extraction allows for detailed spatial analysis. Similarly, GeoChat \cite{geochat} builds on a CLIP-ViT (L-14) backbone, enhancing fine-grained object recognition by upsampling positional encodings to higher resolutions, improving its ability to detect small-scale objects and subtle variations in remote sensing imagery.

For \textbf{text encoding}, most conversational VLMs utilize pre-trained LLMs such as LLaMA \cite{LLaMA} and Vicuna \cite{geochat}. These models process both visual features (converted into textual representations) and user queries to generate structured responses. RS-LLaVA \cite{RS-llava} applies Low-Rank Adaptation (LoRA) \cite{LoRA} to fine-tune LLaMA for remote sensing-specific applications, balancing computational efficiency with task adaptability.

Another essential aspect of conversational VLMs is their \textbf{vision-language alignment mechanism}. Many models rely on MLP-based projection layers, such as SkyEyeGPT \cite{skyeyegpt}, which employs a single-layer MLP, while others like RS-LLaVA use a dual-layer projection. More sophisticated approaches explore learnable query embeddings for structured reasoning, such as LHRS-Bot~\cite{Lhrs-bot}, which introduces hierarchical queries to extract multilevel semantic representations from images. 

\section{VLMs for EO Regression: Problem Setting}
\subsection{Regression Problem Definition}

Consider a standard regression problem setting in which we have a training dataset:
\[
\mathcal{D} = \{(\mathbf{x}_i, y_i) \mid i = 1, 2, \ldots, N\},
\]
where:
\begin{itemize}
    \item \(\mathbf{x}_i \in \mathcal{X}\) denotes the input (e.g., images),
    \item \(y_i \in \mathbb{R}\) denotes the target output (e.g., a continuous values to be estimated).
\end{itemize}

The goal is to learn a function \( f_{\boldsymbol{\theta}} : \mathcal{X} \to \mathbb{R} \) parameterized by \(\boldsymbol{\theta}\). For any new (unseen) sample \(\mathbf{x}\), we wish to predict the corresponding continuous value \(\hat{y}\). Typically, we achieve this by minimizing a loss function. A common choice is the Mean Squared Error (MSE), leading to the following objective:
\[
\boldsymbol{\theta}^* = 
\underset{\boldsymbol{\theta}}{\mathrm{argmin}}
\ \mathcal{L}(\boldsymbol{\theta}) 
= 
\underset{\boldsymbol{\theta}}{\mathrm{argmin}}
\ \frac{1}{N}\sum_{i=1}^N \bigl(f_{\boldsymbol{\theta}}(\mathbf{x}_i) - y_i\bigr)^2.
\]

In the context of EO tasks, \(\mathbf{x}_i\) typically consists of EO data, such as multispectral, SAR, or optical imagery, while \(y_i\) represents a geophysical or environmental variable (e.g., temperature, humidity, PM2.5 concentration, or vegetation index). 

When extending this framework to \textit{VLMs for Regression}, the input \(\mathbf{x}_i\) becomes inherently multi-modal, incorporating not only visual data but also auxiliary textual descriptions, temporal sequences, or structured metadata. The output \(y_i\) remains a continuous target variable, but the optimization process must now accommodate the fusion of heterogeneous modalities. The primary objective remains to learn a mapping function \(f_{\boldsymbol{\theta}}\) that effectively integrates multi-modal information to predict \(y_i\) with high accuracy in a regression setting.

\begin{table*}[htbp]
\centering
\small
\caption{Comparison of Regression and Perception \& Description tasks in Earth Observation.}
\renewcommand{\arraystretch}{1.2}
\begin{tabular}{p{3.5cm} p{6.2cm} p{7.2cm}}
\toprule
\textbf{Aspect} & \textbf{Regression} & \textbf{Perception \& Description } \\
\midrule

\textbf{Output Type} & 
\textbf{Continuous values} (e.g., temperature). 
 &  Primarily \textbf{classes}, \textbf{ boxes}, \textbf{masks}, \textbf{texts}. \\
\textbf{Characteristics} & 
Requires \textbf{precise quantitative estimation}. & 
Emphasizes \textbf{ visual content semantic understanding}. \\

\textbf{Evaluation Metrics} & 
\textbf{MSE}, \textbf{MAE}, \textbf{RMSE}, \textbf{R²}, etc. 
&\textbf{Acc}, \textbf{mAP}, \textbf{IoU}, \textbf{F1},\textbf{BLEU}, \textbf{METEOR}, etc. \\

\bottomrule
\end{tabular}

\label{tab:regression_vs_perception}
\end{table*}

\begin{table*}[htbp]
\centering
\caption{Regression in image understanding vs.Regression in EO scientific problem.}
\small
\renewcommand{\arraystretch}{1.5}
\begin{tabular}{p{3.8cm} p{6.5cm} p{6.5cm}}
\toprule
\textbf{Aspect} & \textbf{Regression in image understanding} & \textbf{ Regression in EO scientific problem} \\
\midrule

\textbf{Focus} & 
Exploration and reconstruction of \textbf{visual geometry}. & 
Accurate modeling of \textbf{ecological, geophysical, and physical phenomena}. \\

\textbf{Key Objective} & 
Ensure \textbf{visual consistency} in predictions (e.g., smooth depth maps, reasonable geometric structures). & 
Align predictions with \textbf{real-world geophysical values} (e.g., soil moisture, NDVI), rather than visual attributes. \\

\textbf{Relation to Visual Features} & 
Directly related to \textbf{visual feature extraction} (e.g., texture, spatial relationships, lighting). & 
Primarily relies on data features but may not directly correlate with the image’s \textbf{visual appearance}. \\

\textbf{Uncertainty Control} & 
Less critical but relevant for safety in applications like robotics and autonomous systems. & 
High emphasis on \textbf{uncertainty estimation} to ensure reliable scientific and operational decision-making. \\


\bottomrule
\end{tabular}

\label{tab:depth_vs_eo}
\end{table*}

\subsection{Scientific Regression vs. Perception Generation in EO}

Scientific regression and perception generation represent two distinct approaches in EO data interpretation (see Table \ref{tab:regression_vs_perception}). Current VLMs always rely on token-by-token generation for perception tasks,such as image captioning \cite{xu2024pllava}, which focuses on interpreting observable features to produce discrete outputs. In contrast, scientific regression tasks, like biomass prediction or the estimation of ecological indicators, require the continuous generation of numerical values that accurately reflect underlying physical processes.

This fundamental difference in output modalities not only underscores the limitations of applying a unified VLM architecture to both types of tasks but also highlights the need for specialized model adaptations. While existing VLMs excel at perception-oriented tasks through discrete token generation, the continuous nature of scientific regression demands alternative strategies that extend beyond visual recognition to robustly reconstruct real-world parameters. Addressing this gap is essential for achieving accurate quantitative estimates in EO applications.

Scientific regression and perception generation are two pivotal facets of EO data interpretation (see Table \ref{tab:regression_vs_perception}). Current Vision-Language Models (VLMs) typically employ token-by-token generation for perception tasks, such as image captioning \cite{xu2024pllava}, which relies on the direct interpretation of observable features within an image. In contrast, scientific regression tasks, for instance in biomass prediction or the estimation of ecological indicators, necessitate the ability to generate continuous numerical values that accurately reflect underlying physical processes. This fundamental discrepancy in output generation has thus far precluded the development of a unified VLM model capable of handling both regression and generation tasks effectively.

This section aims to clarify these differences and emphasizes the need for specialized model adaptations to achieve accurate quantitative estimates in EO regression tasks. While VLMs have demonstrated strong performance in discrete token generation for perception-oriented tasks, the continuous nature of regression outputs calls for alternative strategies that extend beyond visual recognition to reconstruct parameters tied to real-world physical phenomena.

\begin{figure}
    \centering
    \includegraphics[width=1.0\linewidth]{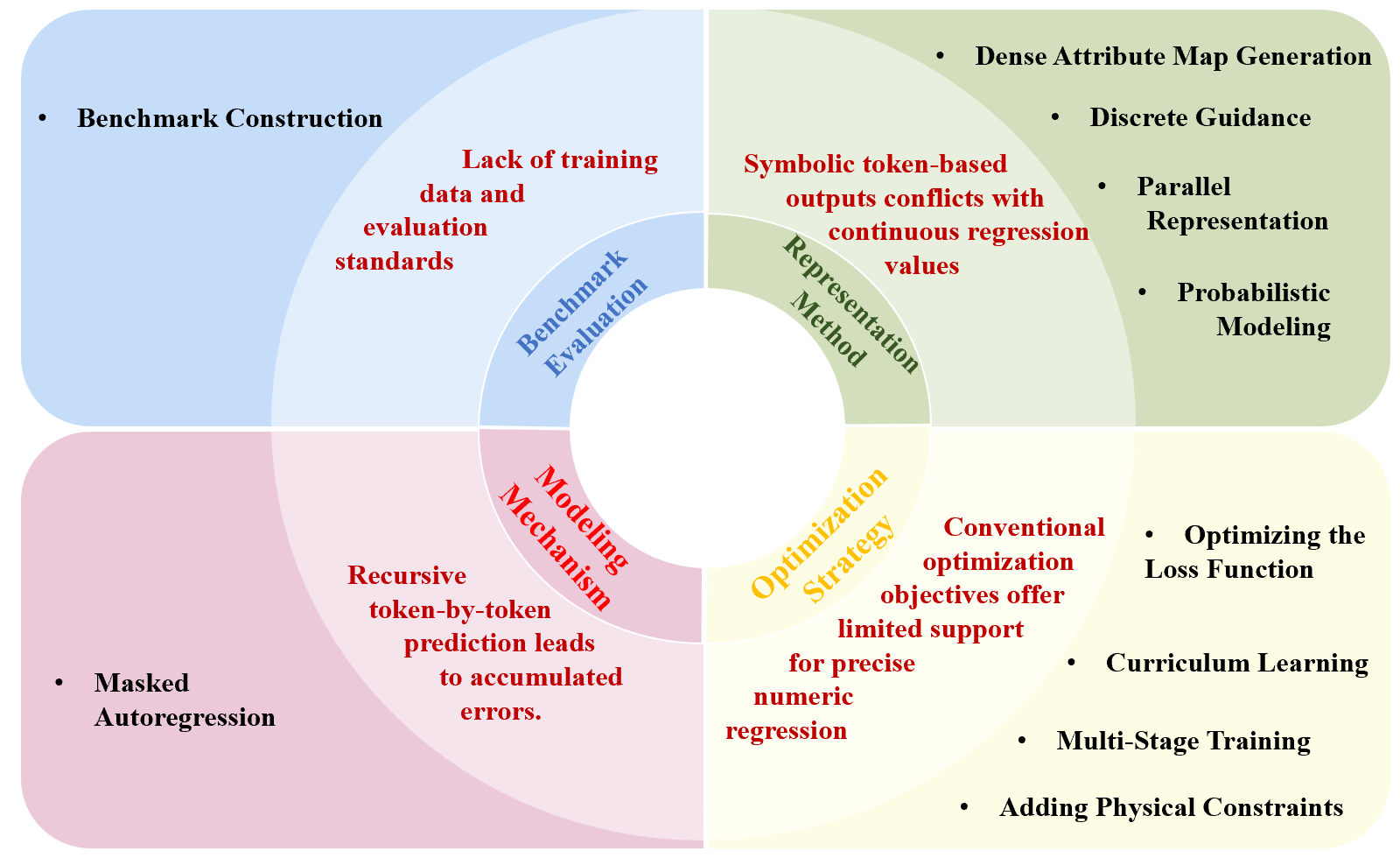}
    \caption{Challenges \& Methodological insights in VLMs for EO Regression.}
    \label{fig:challenges}
    \vspace{-2mm}
\end{figure}

\subsection{Regression in Image Understanding vs. Regression in EO Scientific Problems}

Table~\ref{tab:depth_vs_eo} summarizes the key distinctions between regression tasks in image understanding and those in EO scientific problems. In image understanding tasks, regression primarily focuses on the exploration and reconstruction of visual geometry. The key objective here is to ensure visual consistency, for instance by producing smooth depth maps or maintaining coherent geometric structures. This process relies directly on extracting and interpreting visual features such as texture, spatial relationships, and lighting conditions.

In contrast, regression tasks within EO scientific problems are designed to accurately model complex ecological, geophysical, and physical phenomena. Rather than emphasizing visual consistency, the main goal is to align predictions with real-world geophysical values, such as soil moisture levels or NDVI (Normalized Difference Vegetation Index). This type of regression, although it may utilize data derived from images, does not always correlate directly with the image's visual appearance. Instead, it depends on numerical data that represent underlying physical processes.

Another critical difference highlighted in Table~\ref{tab:depth_vs_eo} is the treatment of uncertainty. While uncertainty control in image understanding is relevant, especially for safety-critical applications like robotics or autonomous systems, it is of paramount importance in EO scientific tasks. Here, high-precision uncertainty estimation is essential to ensure reliable scientific analyses and informed operational decision-making.

Overall, this comparison underlines that while both domains employ regression, the nature of the output and the underlying objectives necessitate distinct approaches. Image understanding regression is closely tied to visual features and spatial consistency, whereas EO scientific regression requires a robust framework for continuous numerical prediction and rigorous uncertainty control.

\section{VLMs for EO Regression: Challenges}
\label{sec:challenges}

Although recent advances in VLMs provide new avenues for EO regression tasks, several significant challenges must be addressed. Below, we discuss four key obstacles.

\paragraph{\textbf{Benchmark: Lack of training data and evaluation standards}}
VLM-based EO regression often suffers from a scarcity of large, high-quality datasets. Unlike well-established language or vision tasks with benchmarks like GLUE~\cite{wang2018glue}, ImageNet~\cite{deng2009imagenet} and LAION2B~\cite{schuhmann2022laion}, EO regression lacks standardized datasets that capture the full diversity of geospatial conditions and sensor modalities. This shortage complicates model training and validation, as researchers have no widely accepted baseline for gauging performance. Moreover, the absence of domain-specific metrics for regression, beyond generic measures~\cite{hodson2022root} like MAE (Mean Absolute Error) or RMSE (Root Mean Squared Error), makes it difficult to comprehensively evaluate the models’ effectiveness in real-world applications, such as climate monitoring and agricultural yield prediction.

\paragraph{\textbf{Representation method: Conflict between discrete symbolic representation and continuous value prediction}}
VLMs typically treat both textual and numerical tokens as discrete symbols, optimizing the next-token prediction likelihood. While this approach works well for language-related tasks, it may be at odds with continuous value prediction in EO regression. Subtle numerical differences in reflectance or temperature can be lost when quantized into symbolic forms, reducing the precision needed for accurate modeling of physical processes. The discrete nature of token-based representations can also limit the model’s ability to interpolate or extrapolate smoothly, a critical capability in regression tasks where data often spans a continuous range of values.

\paragraph{\textbf{Modeling mechanism: Amplified cumulative error from next-token prediction}}
In standard language generation, small token-level errors are often tolerable because semantically similar words can be substituted with minimal impact on overall meaning. For regression tasks, however, inaccuracies compound more severely. Each erroneous token in a generative sequence can shift the subsequent numerical predictions, leading to significant deviations from true values by the end of the output. This issue is exacerbated by EO data’s sensitivity to subtle changes in reflectance or temperature measurements, where even minor cumulative errors can lead to large discrepancies in downstream analyses.

\paragraph{\textbf{Optimization strategy: Lack of direct numerical optimization}}
The training objective of most VLMs is built around maximizing the likelihood of generating correct tokens, rather than explicitly minimizing numerical errors. This approach can neglect the fine-grained numerical calibration required by EO regression tasks, where accurate estimation of continuous variables is paramount. Although techniques like reinforcement learning from human feedback or tailored loss functions can partially address this gap, they are less mature in the context of token-based generative architectures. As a result, VLMs may converge to solutions that perform well in a symbolic, lexical sense yet fail to produce precise regression outputs essential for practical EO applications.

\begin{figure*}
    \centering
    \includegraphics[width=1.0\linewidth]{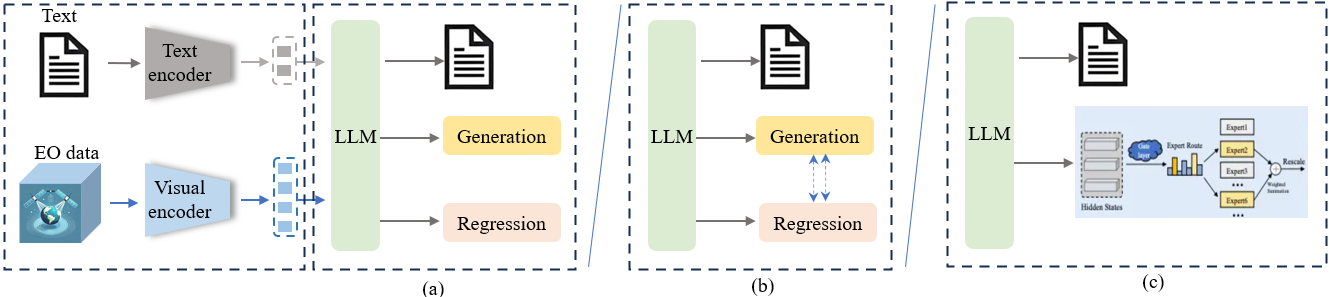}
\caption{Parallel representation: separating discrete and
continuous prediction. (a) Independent regression head architecture. (b) Information-sharing dual-head architecture. (c) MOE based based dual-head architecture}
    \label{fig:Parallel}
    \vspace{-2mm}
\end{figure*}

\section{VLMs for EO Regression: Methodological Insights}


In this section, how to handle four major challenges identified in previous analyses are discussed from different perspective (see Fig.~\ref{fig:challenges}). In particular, our discussion is based on the widely adopted open-source LLaVA structure, which serves as a practical example for these approaches. First, we discuss the factors that need to be considered during benchmark construction. Next, to address the conflict between discrete representation and continuous numerical prediction, we initially propose four solutions: \textit{Parallel Representation},  \textit{Discrete Guidance}, \textit{Dense Attribute Map Generation} and \textit{Probabilistic Modeling}. Then, we introduce a \textit{masked autoregression approach} to mitigate the cumulative error amplification associated with the VLM modeling mechanism. Additionally, we provide several potential solutions to overcome the challenges arising from the lack of  \textit{numerical optimization}. Finally, we delve into the potential of other popular techniques in the field, such as  \textit{distillation and prompt tuning}, for addressing EO regression problems.


\subsection{Benchmark construction}

In the field of EO, most existing datasets are tailored towards image perception and description tasks, encapsulated in a "see and describe" paradigm. However, regression problems within EO involve a "see and analyze" approach, where the inference targets are not directly observable from the image content but are intrinsic attributes of the Earth's surface. A robust benchmark~\cite{xue2024reo} is essential to advance regression research in EO using VLMs. The following considerations are critical in constructing such a benchmark:
\begin{itemize}
    \item \textbf{Global spatial distribution.} EO tasks necessitate datasets with globally representative samples. The spatial distribution of the data must be balanced to ensure the dataset accurately reflects diverse geographic and environmental conditions worldwide.
    \item \textbf{Sufficient and necessary data sources.} Multi-sensor data integration is vital. The data must originate from an array of sensors to provide a comprehensive foundation for regression analysis, yet should avoid redundancy. Including unnecessary sensors increases computational overhead without adding value to scientific insights.
    \item \textbf{Integration of relevant scientific knowledge.} Text annotations in the dataset should embed measurable scientific indicators relevant to the inference tasks. Beyond descriptive metadata of the imagery, these annotations must bridge the gap between visual data and the underlying scientific phenomena. This integration enhances the model's ability to infer latent variables, effectively addressing the information bottleneck inherent in image-based reasoning.
     \item \textbf{Temporal coverage and evolutionary trends.} Many EO tasks require the detection and analysis of changes over time (e.g., vegetation cycles, climate trends, or seasonal variations). A well-constructed benchmark must capture multi-temporal data spanning suitable time intervals. By integrating such temporal information, the dataset allows VLMs to model dynamic processes more accurately, enabling deeper insights into long-term environmental patterns and improving the reliability of regression outcomes for trend-focused scientific inquiries.
    \item \textbf{Semantic sensitivity of models.} Regression tasks require models to prioritize semantic content in text over superficial patterns. The benchmark should assess whether models are sensitive to meaningful scientific semantics rather than stylistic elements or filler words.

\end{itemize}

By considering these factors, the constructed benchmark will support the development of VLMs that excel in regression tasks, ultimately enhancing our understanding of complex environmental systems through EO data.

\subsection{Parallel Representation}
One intuitive approach as illustrated in Fig.~\ref{fig:Parallel}(a) is to introduce an additional regression head alongside the original text-generation head. Both heads receive the same input (text and images) but operate separately. This design is simple and ensures that discrete and continuous tasks do not interfere with each other. However, a notable \emph{drawback} is that there is \textbf{no information sharing} between the two heads: the numerical regression head cannot take advantage of the intermediate knowledge from the text-generation process.


To address this issue, the model shown in Fig.~\ref{fig:Parallel}(b) incorporates \textbf{information-sharing mechanisms} between the text-generation and regression heads, such as cross-attention or gated feature fusion, enabling the numerical prediction to utilize linguistic insights. A more \emph{flexible} architecture in Fig.~\ref{fig:Parallel}(c) introduces a Mixture of Experts (MoE) mechanism~\cite{lin2024moe}, where the LLM output hidden state passes through a gate layer that selects among parallel expert networks. Some experts may be specialized in continuous regression tasks, while others excel at text-based reasoning. The final prediction is a weighted aggregation of the selected experts’ outputs, allowing the model to \textbf{dynamically allocate} different aspects of the input to the most suitable experts. While this approach can significantly \textbf{enrich the representational capacity} and \textbf{share prior knowledge} across experts, it also entails higher training complexity and may be more prone to instability if not carefully regularized and tuned.

In summary, these parallel representation methods offer a flexible way to handle discrete and continuous signals simultaneously. The introduction of MoE, in particular, extends the architecture’s expressiveness by letting the model learn which experts to activate for each input, thereby improving performance for both regression and text-generation tasks. However, additional hyper-parameter tuning and more sophisticated optimization strategies may be required to maintain stability and scalability. 


\begin{figure}
    \centering
    \includegraphics[width=1.0\linewidth]{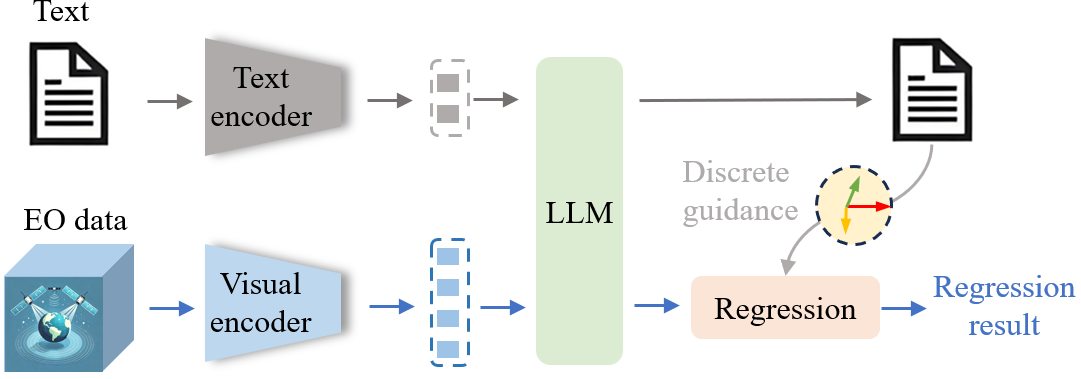}
    \caption{Discrete-guided continuous prediction.}
    \label{fig:Discrete-guided}
    \vspace{-2mm}
\end{figure}
\subsection{Discrete Guidance}

A promising solution to address the challenge of reconciling discrete representations with continuous numerical prediction is to discretize continuous values into intervals and leverage these discrete predictions to guide the continuous regression process. Specifically, as illustrated in Fig.~\ref{fig:Discrete-guided}, the VLM retains a \emph{text-generation head}, but when predicting a continuous value, this head first outputs a \textbf{range or interval} for the number (e.g., ``the value is high, likely in the range [100, 200]''). That interval is then \emph{softly} integrated into a regression head via mechanisms such as cross-attention, convolution, or other fusion methods. By doing so, the regression head can benefit from the linguistic priors learned by the model, which is particularly advantageous when the numerical scale is large or highly variable.

The main drawback of this approach lies in its \textbf{dependence on the design of the intervals}. If the intervals are too coarse, numerical precision will be limited. Conversely, if they are too fine, the classification complexity increases, potentially necessitating domain-specific knowledge to define the intervals appropriately.



\subsection{Dense Attribute Map Generation}

\begin{figure}
    \centering
    \includegraphics[width=1.0\linewidth]{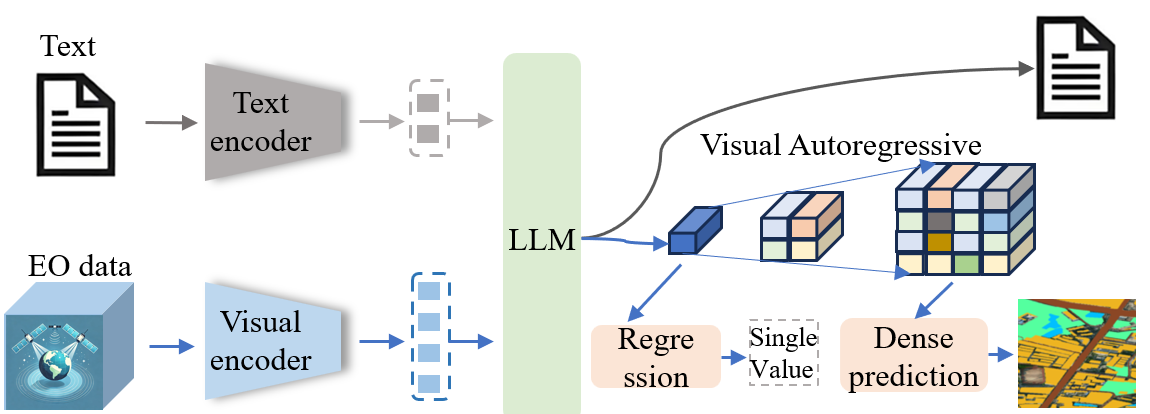}
    \caption{Transforming dense regression into attribute map generation.}
    \label{fig:VAR}
    \vspace{-2mm}
\end{figure}

A novel perspective on dense regression task EO, where each pixel predicts a continuous value representing a physical attribute (e.g., temperature, precipitation, or vegetation index) is to \emph{reinterpret} this task as an attribute map-generation problem within a VLM framework. Specifically, if one treats the pixel-wise target values as a projection on a latent physical dimension, generating a high-fidelity “attribute map” in this dimension is akin to reconstructing a scene from continuous signals. In practice, the VLM receives both textual inputs (e.g., location descriptions, temporal constraints, scientific parameters) and the original EO imagery; it then produces a fused embedding. Instead of decoding this embedding solely into text or discrete class labels, we attach a generative head that outputs a 2D map of the desired physical quantity, as shown in Fig.~\ref{fig:VAR}, The textual context from the VLM can provide guiding constraints or domain knowledge (e.g., known temperature ranges, geophysical phenomena), while the vision component ensures the spatial fidelity to the input image content.

Two main strategies can be considered to implement this generative decoding~\cite{chen2025janus}. The first is \textbf{Generation via Diffusion Models}~\cite{zhou2024transfusion, xie2024show}: the fused embedding from the VLM can condition a diffusion network that denoises random latent tensors into detailed “physical attribute maps.” Diffusion-based methods excel at capturing complex data distributions, although they may be computationally expensive and slow at inference time for high-resolution outputs. The second is \textbf{Generation via Codebooks} \cite{team2024chameleon}, where the latent space is discretized into a finite codebook of embeddings, and the VLM directly predicts which code vectors correspond to each spatial location auto-regressively. This discrete approach eases the control and potentially speeds up generation, but may introduce quantization errors when the codebook size is insufficient to represent subtle continuous variations. 

By integrating textual, visual, and generative components within a single VLM pipeline, we can harness textual prompts, metadata, and physical constraints to steer the generative head toward producing high-quality reconstructions in the target attribute variable space. This integrated approach produces outputs that are visually interpretable, which facilitates rapid qualitative assessment, and it enriches the results with scientific context provided by language, a feature that is particularly valuable for specialized EO domains.

Nonetheless, matching the purely “visual” representation of the generative output to real-world physical scales remains a challenge.  Diffusion-based methods require substantial computing resources and longer inference times, whereas codebook-based methods risk losing nuances in continuous signals. Despite these trade-offs, converting dense regression to generative modeling presents a promising avenue for tasks where interpretability, controllability, or large-scale data generation are central requirements.

\begin{figure*}
    \centering
    \includegraphics[width=1.0\linewidth]{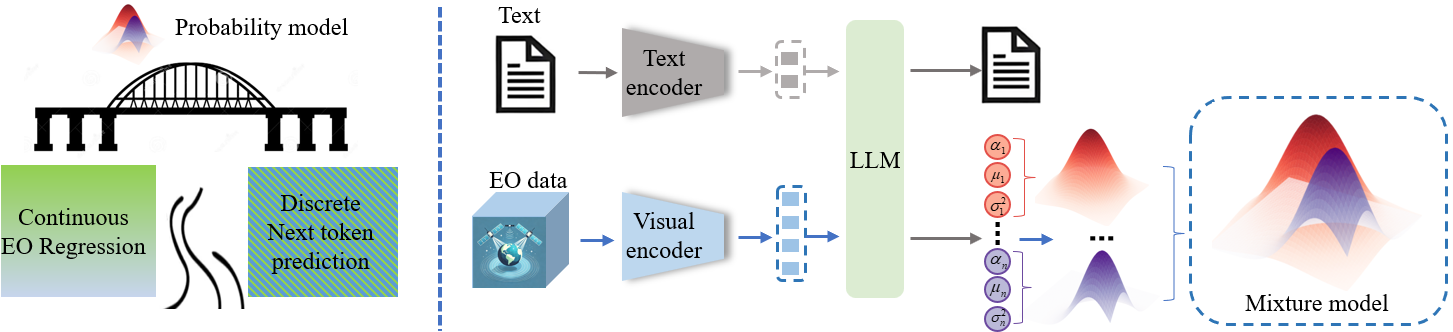}
    \caption{Bridging discrete and continuous with a mixture of gaussian modeling.}
    \label{fig:MDN}
    \vspace{-2mm}
\end{figure*}

\subsection{Probabilistic Modeling}

In this approach, the VLM estimates parameters of a Mixture of Gaussians distribution~\cite{bishop1994mixture,guillaumes2017mixture} to perform numerical regression. The overall process can be outlined as follows:

\begin{enumerate}
    \item \textbf{Input}: The model extract image features $f_{\mathrm{img}}$ and text features $f_{\mathrm{text}}$, combining them (e.g., via attention) into a single hidden representation $h$.
    \item \textbf{Parameter Prediction}: From $h$, the model predicts the mixture parameters, including mean $\mu_k$, variance $\sigma_k$, and mixing coefficient $\pi_k$ for each Gaussian component ($k=1,\dots,K$):
    \[
    \{\mu_k, \sigma_k, \pi_k\}_{k=1}^K = \mathrm{MLP}(h).
    \]
    \item \textbf{Distribution Computation}: The continuous distribution is thus:
    \[
    p(x) = \sum_{k=1}^K \pi_k \mathcal{N}\bigl(x \mid \mu_k, \sigma_k^2\bigr),
    \]
    where $K$ is the predefined number of Gaussian components.
    \item \textbf{Regression Output}: The predicted value can be taken as the expected value of this mixture:
    \[
    \hat{x} = \sum_{k=1}^K \pi_k \mu_k,
    \]
    and \textbf{uncertainty}~\cite{gawlikowski2023survey} can be similarly estimated by computing the variance:
    \[
    \mathrm{Var}(x) = \sum_{k=1}^K \pi_k \bigl(\sigma_k^2 + \mu_k^2\bigr) - \hat{x}^2.
    \]
\end{enumerate}

This probabilistic strategy offers flexible distribution modeling suitable for multimodal or multi-peaked data while naturally supporting uncertainty estimation, and it can integrate the VLM’s priors and image cues into the mixture parameters for potentially better fits to complex data distributions, as shown in Fig.~\ref{fig:MDN}. however, the training process may be unstable due to gradient issues or component collapse, the choice of $K$ heavily depends on domain expertise, and large numbers of parameters demand careful regularization and early stopping.

\subsection{Masked Autoregression}

\begin{figure}
    \centering
    \includegraphics[width=1.0\linewidth]{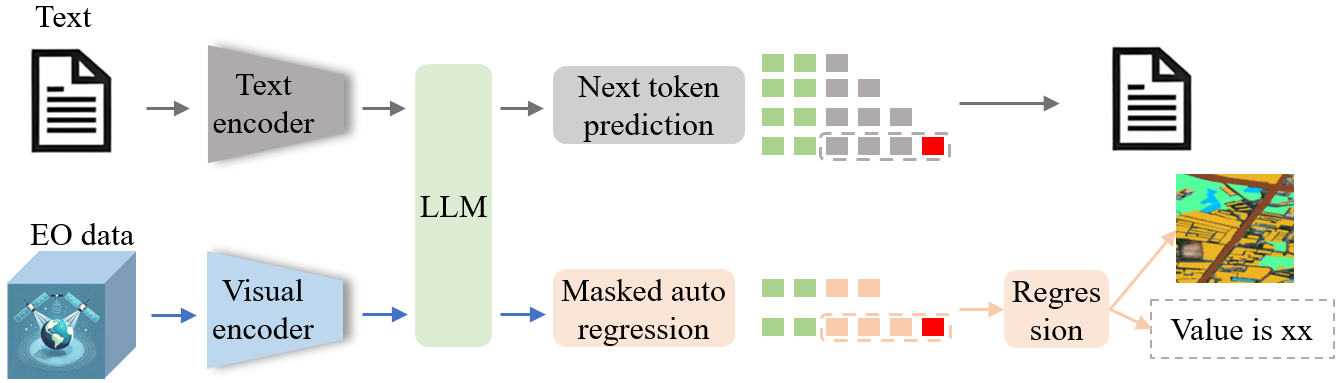}
    \caption{Masked autoregression for dense regression.}
    \label{fig:MAR-regression}
    \vspace{-2mm}
\end{figure}

Building on the image-generation approach discussed earlier, we explore to adopt a masked autoregressive strategy~\cite{chang2022maskgit} within the image branch to mitigate cumulative error in dense regression tasks, as shown in Fig.~\ref{fig:MAR-regression}. Instead of predicting the entire spatial map in one pass, the model iteratively refines its output through a mask-and-refill process. In each iteration, a subset of pixel values is masked, and the model fills in these positions conditioned on the unmasked tokens. This iterative process allows the model to correct early mistakes rather than allowing errors to propagate unchecked through the entire prediction chain.

The core reason that this masked autoregressive strategy helps reduce cumulative error lies in its ability to break the strict sequential dependency inherent in traditional autoregressive methods. In conventional autoregressive prediction, each new pixel value is generated based on all previous outputs. If an early prediction error occurs, it is incorporated as part of the context for all subsequent predictions, leading to a snowball effect. By contrast, the masked autoregressive approach periodically resets the prediction context. During each mask-and-refill cycle, only a portion of the output remains fixed while the rest is re-estimated. This not only limits the influence of early errors but also provides an opportunity for the model to leverage global contextual information from the unmasked regions to correct or mitigate those errors.

Furthermore, the iterative nature of the process introduces a form of self-correction. As more iterations are performed, the prediction gradually converges toward a consistent and coherent reconstruction of the target spatial map. Each iteration refines the previously generated outputs based on both local details and global context, reducing discrepancies between the prediction and the true underlying physical attributes. The masked regions, being predicted later in the process, are less likely to be tainted by earlier errors since they benefit from the progressively improved context established in prior rounds.

This approach also benefits from the integration of cross-modal information. The text-generation branch continues to provide rich domain-specific guidance through language, ensuring that the iterative refinement process is not solely reliant on visual features. The model can thereby use textual cues to impose physical constraints and rectify potential deviations introduced in earlier iterations.

Overall, by allowing the model to re-evaluate and adjust its predictions at multiple stages, the masked autoregressive strategy significantly reduces the impact of any single error, leading to more robust and accurate reconstructions in dense EO regression tasks.

\begin{figure*}
\centering
    \includegraphics[width=1.0\linewidth]{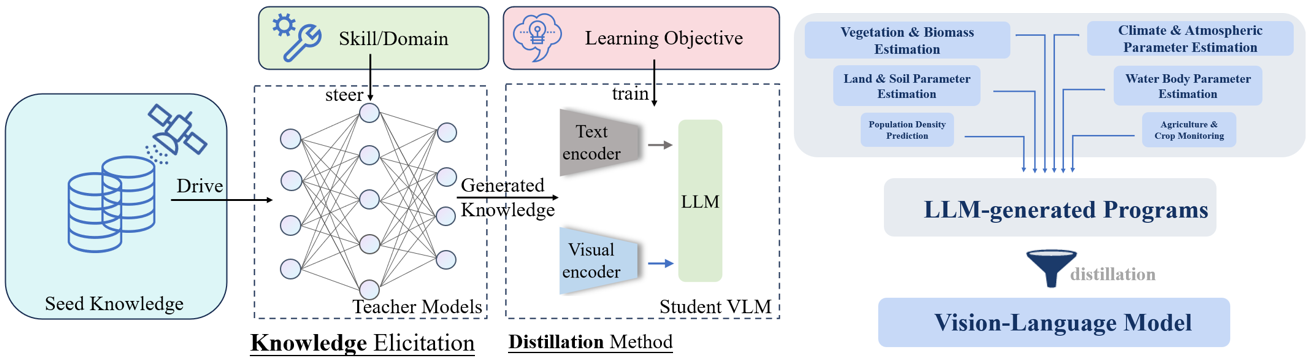}
    \caption{Distillation and ensemble method.}
    \label{fig:distillation}
    \vspace{-2mm}
\end{figure*}

\subsection{Model Optimization Strategies}

Robustly optimizing a VLM for regression tasks in EO requires techniques that go beyond straightforward parameter updates. Here, we explore several key strategies: \emph{designing loss functions} that accurately capture continuous-valued errors, \emph{integrating physical constraints} to reflect domain knowledge, \emph{curriculum learning} to ease training complexity, and \emph{multi-stage training} approaches that progressively refine model performance.

\noindent
\textbf{Optimizing the Loss Function.}  
To capture the continuous nature of EO regression, the most direct approach is to use standard objectives such as Mean Squared Error (MSE) or Mean Absolute Error (MAE). MSE emphasizes penalizing larger deviations more strongly, which can be beneficial for tasks where large errors are unacceptable (e.g., estimating extreme climatic events). In contrast, MAE is more robust to outliers but may underemphasize substantial discrepancies. Beyond these conventional metrics, domain-specific losses may incorporate statistical or physical properties of interest, such as penalizing phase-shifted predictions differently or emphasizing spatially correlated structures. Selecting a proper balance among these losses is crucial for stable and interpretable training outcomes.

\noindent
\textbf{Adding Physical Constraints.}  
Many EO problems entail \emph{a priori} knowledge about the quantity being predicted, such as a known feasible range (e.g., temperature bounds) or conservation laws (e.g., water or energy balance). Incorporating these constraints directly into the model or loss function can reduce unphysical predictions and guide the training process toward meaningful solutions. This can be achieved by imposing \emph{penalty terms} for violating known physical relationships, or by enforcing soft or hard bounds on outputs. For instance, a temperature estimate could be clamped within plausible extremes, or a divergence-free constraint might be applied for certain fluid-flow problems. By intertwining domain expertise with learned representations, the model gains better generalization in realistic scenarios.

\noindent
\textbf{Curriculum Learning.}  
When the training data contain diverse conditions ranging from trivial (e.g., clear-sky imagery) to highly complex (e.g., heavy cloud cover or sensor noise), the model can struggle to converge if all samples are treated equally from the start. Curriculum learning~\cite{wang2021survey} addresses this by presenting examples in order of increasing difficulty, allowing the network to master simpler tasks before tackling more intricate ones. For regression in EO, difficulty can be defined by noise levels, data quality, or the complexity of the underlying physical processes. A gradual exposure strategy often stabilizes training, leading to smoother loss curves and potentially higher accuracy on challenging scenarios.

\noindent
\textbf{Multi-Stage Training.}  
Building on the idea of curriculum learning, \emph{multi-stage training} breaks the optimization into sequential phases, each focusing on specific objectives or subsets of data. For instance, an initial stage could refine basic visual-linguistic alignment and medium-level feature extraction, ensuring the model accurately interprets input imagery and textual context. Subsequent stages could zoom in on fine-grained regression for narrower domains, add specialized constraints (such as physical priors), or incorporate novel data augmentations. By progressively narrowing the objective, the model internalizes foundational knowledge first, then specializes in a controlled manner, reducing catastrophic forgetting and facilitating more robust convergence.

Overall, these optimization strategies provide a systematic way to enhance VLMs’ capacity to handle complex, continuous-valued tasks in EO. The interplay of carefully designed objectives and informed training schedules empowers the model to achieve both scientific plausibility and strong predictive performance. 


\subsection{Distillation and Ensemble Methods}
As illustrated in Fig.~\ref{fig:distillation}, to enhance regression performance in EO tasks using VLMs, one can distill ~\cite{xu2024survey} knowledge from an existing \emph{teacher model} (e.g., a well-calibrated physical model or a traditional ML regressor) into a \emph{student model} (the VLM). The teacher’s outputs on the training data act as “soft labels” or reference targets for the student. During inference, an ensemble of several models (including the distilled student) can be leveraged to obtain more robust predictions.

\medskip
\noindent
\textbf{Knowledge Distillation.}\quad 
In classical knowledge distillation, let \(p_t(y \mid x)\) denote the teacher’s prediction for input \(x\), and \(p_s(y \mid x)\) the student’s. 
For regression, we can treat \(p_t\) and \(p_s\) either as continuous density estimates or direct numeric outputs. A common distillation loss integrates both the \emph{teacher’s prediction} and the \emph{ground truth}, for instance:
\[
\mathcal{L}_{\text{distill}} 
= \sum_{(x,y)\in D} 
\Bigl(
  \alpha \,\|y_s - y_t\|^2 + \beta \,\|y_s - y_{\mathrm{gt}}\|^2
\Bigr),
\]
where \(y_s\) is the student’s prediction, \(y_t\) is the teacher’s prediction, \(y_{\mathrm{gt}}\) is the ground-truth label, and \(\alpha\), \(\beta\) control the relative weighting of distillation and standard supervised loss. This allows the student VLM to inherit strong domain priors (e.g., from a physics-based model) and align its output distribution with both the teacher’s estimates and real-world measurements. Although this helps jump-start the learning process, the student’s performance is bounded by the quality and limitations of the teacher model.

\medskip
\noindent
\textbf{Ensemble Inference.}\quad 
Once trained, the student VLM can be paired with other models to form an ensemble. Suppose we have \(M\) different models that output predictions \(\hat{y}_1, \hat{y}_2, \ldots, \hat{y}_M\) for the same input \(x\). A simple weighted average yields the final output:
\[
y_{\mathrm{ensemble}} 
= \sum_{i=1}^{M} w_i \,\hat{y}_i,
\quad \text{where}~\sum_{i=1}^{M} w_i = 1.
\]
Weights \(\{w_i\}\) can be tuned based on each model’s past performance or uncertainty estimates. More sophisticated fusion methods (e.g., soft or hard voting, or Bayesian uncertainty fusion) may be employed to combine complementary model strengths while mitigating individual weaknesses.

\noindent
\textbf{Input--Output Setup.}\quad 
During training, the \emph{multimodal inputs} include EO imagery (satellite or hyperspectral data) alongside textual metadata or domain descriptions. The teacher model, which might be a robust physical simulator or a traditional ML model (e.g., random forest, XGBoost), provides its predictions \(\{y_t\}\) for each sample. The \emph{student} VLM then outputs \(y_s\) during each training step, and the \emph{distillation loss} enforces alignment with both \(\{y_t\}\) and the ground truth \(\{y_{\mathrm{gt}}\}\). In the \emph{inference stage}, each participating model (the now-distilled VLM plus any additional models) produces a numeric prediction for a given EO input, which is subsequently combined via the ensemble procedure to yield the final regression outcome.

\noindent
\textbf{Optimization Process.}\quad 
\textit{(1) Distillation Training:} The teacher first infers predictions on the training set. The student VLM processes each sample (imagery + text) to produce its own regression estimate and is penalized by the combination of teacher-based MSE (\(\|y_s - y_t\|^2\)) and data-based MSE (\(\|y_s - y_{\mathrm{gt}}\|^2\)). Through gradient descent (e.g., Adam or SGD), the student converges on a parameter space that balances faithfulness to the teacher’s domain knowledge and fidelity to the real-world labels.  
\textit{(2) Inference \& Ensemble:} At test time, new EO data are fed to each model in the ensemble. Their predictions are aggregated (e.g., via weighted average), producing a more robust, error-tolerant final result.

\noindent
\textbf{Advantages and Disadvantages.}\quad 
By integrating a strong teacher model, knowledge distillation allows the student to inherit established domain priors while reducing overfitting risks. In parallel, ensembling multiple models (including the distilled student) can boost overall performance by averaging out individual errors. However, the success of distillation hinges on the teacher’s quality. If the teacher exhibits significant bias, the student may inherit these flaws. Training and inference costs also increase because distillation requires teacher predictions on the entire dataset and ensemble methods necessitate multiple model evaluations during inference. Moreover, forcing the student to closely track the teacher might inhibit the discovery of novel features or patterns not captured by the teacher’s representation. Despite these trade-offs, knowledge distillation and ensemble strategies remain straightforward yet powerful complements to other techniques (e.g., multi-task learning, physical constraints, or prompt engineering), jointly promoting accuracy and robustness in VLM-based EO regression. 

\subsection{Prompt Engineering \& Instruction Tuning}

When deploying VLMs for EO regression tasks, it is not only the core architecture and loss function that matter but also \emph{how} the model is prompted or instructed to conduct its reasoning. Prompt engineering~\cite{lester2021power} and instruction tuning~\cite{liu2024visual} comprise techniques designed to guide VLMs more effectively, leveraging the model’s inherent language-understanding capabilities and adaptively shaping its numerical regression outputs.
\begin{figure}
    \centering
    \includegraphics[width=1.0\linewidth]{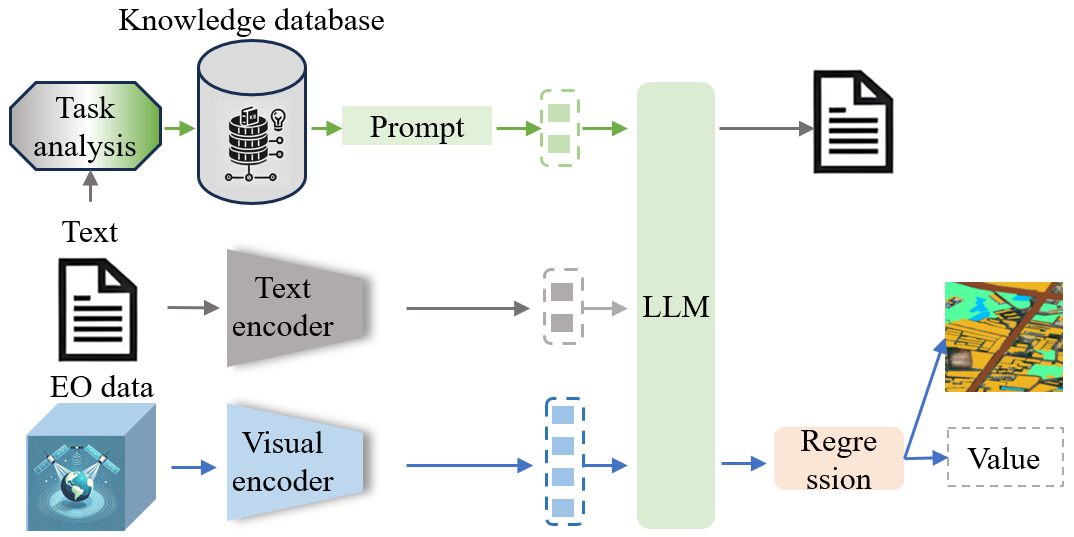}
    \caption{Prompt engineering \& Instruction tuning.}
    \label{fig:prompt-engineering}
    \vspace{-2mm}
\end{figure}

\noindent
\textbf{Prompt Engineering.}\quad 
In typical multimodal frameworks, the text encoder or language component of the VLM receives textual prompts describing the EO scene or specifying the regression goal. As illustrated in Fig.~\ref{fig:prompt-engineering}, by carefully crafting these prompts, practitioners can encourage the model to focus on relevant scientific details, domain constraints, or contextual factors (e.g., meteorological conditions, temporal information, or sensor metadata). For instance, a prompt that emphasizes seasonal changes, known spatial ranges, or physical plausibility (e.g., “predict temperature in the plausible range of 200--320 K”) steers the model toward more reliable outputs. Effective prompt engineering often draws on domain expertise to include key terms or data-related hints, while avoiding ambiguity or irrelevant detail. In some cases, providing negative prompts (i.e., specifying which types of outputs or anomalies are \emph{not} expected) can also help the model refine its predictions.

\noindent
\textbf{Instruction Tuning.}\quad 
Beyond ad-hoc prompt design, \emph{instruction tuning} aims to systematically align the VLM’s behavior with human-defined guidance. Training data are augmented with explicit instructions, such as “Predict the long-term vegetation index trend from this satellite image and text description,” followed by the model’s ideal output. Over a large set of instruction-example pairs, the VLM iteratively learns to parse and follow complex directives, effectively gaining the ability to handle more diverse or specialized queries. In EO regression, instruction tuning can specify not just \emph{what} to predict (e.g., soil moisture, particulate matter concentration) but also \emph{how} to handle edge cases (e.g., missing data, cloud cover) or \emph{which} uncertainties to report (e.g., confidence intervals). As a result, instruction-tuned models become more robust to user variability, requiring fewer manual adjustments at inference time.

\noindent
\textbf{Synergy with Other Techniques.}\quad 
Prompt engineering and instruction tuning integrate naturally with other aspects of a VLM-based EO pipeline. For instance, when using \emph{masked autoregression} for dense regression, prompts can indicate the spatial or temporal segments requiring updates or highlight specific domain priors. Similarly, if \emph{physical constraints} or \emph{knowledge distillation} are in play, instructions might help the model reconcile or rank conflicting information. This synergy empowers the VLM to exploit linguistic cues for better numerical performance, effectively bridging high-level domain knowledge and low-level perceptual features.

\noindent
\textbf{Potential Limitations.}\quad 
Although prompt engineering and instruction tuning can considerably improve performance, they also introduce new design complexities. Crafting optimal prompts typically relies on expert insight and iterative refinement. Even with an extensive instruction dataset, the model might struggle with ambiguous or contradictory instructions and could overfit to particular formats if not exposed to diverse scenarios. Moreover, while these methods enhance interpretability by making the model’s reasoning steps more transparent, they do not guarantee that every response aligns with the underlying ground-truth physics or environmental processes. Hence, prompt engineering and instruction tuning are best viewed as complementary strategies within a broader framework of rigorous modeling, domain-informed constraints, and iterative evaluation.

By carefully designing textual prompts or training on labeled instruction data, EO practitioners can finely steer VLMs toward domain-centric thinking and accurate regression outputs. Whether it involves specifying precise physical ranges or guiding multi-step reasoning via instructions, these strategies leverage the linguistic strengths of VLMs to enhance both usability and performance in complex regression scenarios.

\section{VLMs for EO Regression: Pitfalls}

In addition to these obvious and foreseeable challenges, such as those stated in Section 
\ref{sec:challenges},
there are also hidden issues that could potentially hinder the entire method from working effectively. We refer to these hidden issues as pitfalls. Three potential pitfalls are the cumulative impact of generative errors on regression results, conflicting optimization objectives and scale variability across geographical regions. 

\paragraph{\textbf{Dropout operation losing key information}}
Dropout is commonly used in neural network training to mitigate overfitting by randomly deactivating neurons. However, in the context of EO data regression, subtle cues (e.g., minute changes in spectral reflectance or texture) can be crucial for precise predictions. Excessive dropout could inadvertently discard these important signals, leading to an underestimation of fine-grained details. When combined with a generative mechanism that already dilutes certain features into symbolic tokens, this loss of key information 
may significantly harm regression performance.

\paragraph{\textbf{Insufficient preservation of detailed visual information}} Many VLM architectures rely on compact embeddings or high-level feature representations for images, 
abstracting the information of pixels within a certain domain into high-dimensional features.
While this 
works well in semantic understanding, EO regression tasks often capture 
subtle but meaningful variations on the ground (e.g., subtle color differences indicating crop health). If the model’s visual pathway discards these granular features too aggressively, the subsequent regression component may be deprived of crucial information, leading to suboptimal accuracy. Ensuring that detailed visual information remains accessible via multi-scale feature extraction or domain-specific attention can enhance the model's ability to perform accurate EO regression.


\paragraph{\textbf{Scale variability across geographical regions}}
In EO regression tasks, target variables like biomass or temperature can vary widely across geographical regions, sometimes differing by several orders of magnitude. This variability poses a significant challenge for VLMs, which are not inherently designed to adjust for such scale differences. Without mechanisms to address these variations, models risk overfitting to specific regions, failing to generalize, or producing large errors in areas with extreme values. Integrating domain-specific prior knowledge through the language module in VLMs can help models adapt to these scale differences, improving accuracy and robustness across diverse regions.

\section{Future Research Directions}

In this section, we outline several promising directions for future research in VLMs applied to EO.

\paragraph{\textbf{ Integrating multi-sensor data and optimizing result head}}

Future research can enhance VLMs for EO regression through targeted improvements in \textbf{data integration} and \textbf{model architecture}. Integrating diverse data sources can benefit regression accuracy by providing complementary insights that help VLMs capture complex image patterns. MOE architectures also offer flexibility by dynamically allocating resources across modal inputs, improving model adaptability and accuracy in EO regression. In addition, refining regression head design to match EO data characteristics could improve the precision of VLM outputs. Together, these advancements promise to make VLMs more robust and accurate for EO regression tasks.

\paragraph{\textbf{Exploring multi-temporal regression and trend inference capabilities}} It is necessary to study how to expand VLM capabilities for EO regression to capture trends from multi-temporal EO data. Effectively inferring current scientific indicators based on historical EO data requires that models go beyond spatial feature recognition to accurately interpret temporal dynamics. Research in this area could explore innovative approaches for trend inference, enabling VLMs to handle applications such as climate monitoring and ecosystem change assessment, where temporal patterns are very important. By advancing these capabilities, VLMs could provide more precise and context-aware predictions, leveraging past-to-present dynamics to enhance accuracy in multi-temporal EO regression tasks.


\paragraph{\textbf{Employing multi-step scientific regression}}
Many regression tasks in EO have tightly interdependent relationships, where accurate predictions of some indicators strongly depend on the estimation of foundational indicators, rather than solely on image content. This interdependence calls for VLMs with advanced multi-step regression capabilities, enabling them to sequentially process quantitative steps to arrive at accurate predictions. Future research in multi-step numerical reasoning could enable VLMs to tackle complex EO analyses that require layered, step-by-step calculations rather than single-pass predictions. Advancements in this area could lead to more intelligent, problem-solving VLMs capable of sophisticated and accurate regression for complex EO applications.

\paragraph{\textbf{Facing multi-domain numerical regression challenges}}
In EO, numerous scientific regression tasks are interdependent. Exploring the collaborative optimization of regression problems across different domains could yield deeper insights into human impacts on the natural environment. For instance, jointly predicting population density and ecological parameters may enhance our understanding of how human activities affect ecosystems. Relying solely on imagery for such analyses can be limiting, as it may not capture the complexities of these interactions.

However, this interdisciplinary approach presents significant challenges. Integrating knowledge from multiple domains can exponentially increase optimization complexity, such as the 
the varying dimensionality of input data
and the introduction of greater uncertainties. Additionally, conflicting optimization objectives may arise, complicating the development of effective models. Despite these hurdles, pursuing this research direction offers promising opportunities for advancing global ecological forecasting and understanding human-environment interactions.

\paragraph{\textbf{Quantifying uncertainty in VLM regression predictions}}
Measuring uncertainty is essential when VLMs handle regression tasks, as it directly impacts the reliability of predictions in EO applications. This research direction aims to develop techniques for uncertainty quantification in VLM regression outputs, ensuring that models not only provide predictions but also convey confidence levels. Accurately assessing uncertainty is key to building trustworthy models for critical applications, such as disaster response and environmental monitoring.

\section{Conclusion}

This paper has explored the potential of Vision Language Models (VLMs) for regression tasks in Earth Observation (EO), an area where these models have yet to be fully exploited. We identified key challenges in adapting VLMs to EO regression, including gaps in dedicated benchmarks, mismatches between discrete and continuous representations, the risk of cumulative error accumulation, and the limitations of text-centric training objectives for numerical tasks. Despite these obstacles, VLMs hold significant promise for advancing EO regression, particularly by addressing the information bottleneck inherent in traditional image processing techniques.

We have outlined key methodological insights and proposed strategies to address these challenges, but much remains to be explored. Future research should focus on refining VLM approaches for EO, enhancing their robustness, and incorporating multi-step reasoning for more accurate predictions. As VLMs continue to evolve, they hold great promise for improving the precision and interpretability of EO regression, ultimately contributing to more effective environmental monitoring and decision-making.

\bibliographystyle{IEEEtran}
\bibliography{reference}

\appendices



\ifCLASSOPTIONcaptionsoff
  \newpage
\fi

\end{document}